\documentclass[10pt,twocolumn,letterpaper]{article}

\usepackage{cvpr}              % To produce the CAMERA-READY version
\usepackage{mathtools} % in the preamble
\usepackage[most]{tcolorbox}

\usepackage[most]{tcolorbox}
\tcbuselibrary{listings,breakable}
\usepackage{upquote}

\newtcblisting{promptbox}[1]{%
  enhanced,
  breakable,
  listing only,
  title={#1},
  fonttitle=\bfseries,
  colback=gray!3,
  colframe=black!40,
  boxrule=0.5pt,
  arc=1mm,
  left=1mm,
  right=1mm,
  top=1mm,
  bottom=1mm,
  listing options={
    basicstyle=\ttfamily\scriptsize,
    breaklines=true,
    breakatwhitespace=false,
    columns=fullflexible,
    keepspaces=true,
    showstringspaces=false,
    tabsize=2
  }
}
\usepackage[T1]{fontenc}
\usepackage{textcomp}

\definecolor{cvprblue}{rgb}{0.21,0.49,0.74}
\usepackage[pagebackref,breaklinks,colorlinks,allcolors=cvprblue]{hyperref}
\usepackage{multirow}
\def\paperID{*****} % *** Enter the Paper ID here
\def\confName{CVPR}
\def\confYear{2026}

\title{\method{}: A Diagnostic Suite for Visual Concept Transfer on Natural Images}

\author{
{
Zhaonan Li$^{1}$,
Kyle R. Chickering$^{2,\dagger}$,
Bangzheng Li$^{3}$,
Jacob Dineen$^{1}$,
Xiao Ye$^{1}$,
Zhikun Xu$^{1}$,
Shijie Lu$^{1}$
}\\
{
Yuxi Huang$^{1}$,
Ming Shen$^{1}$,
Bach Nguyen$^{1}$,
Jaya Adithya Pavuluri$^{1}$,
Mau Son Nguyen$^{1}$
}\\
{
Sanika Chavan$^{1}$,
Ngoc Minh Thu Le$^{1}$,
Muhao Chen$^{3}$,
Ben Zhou$^{1}$
}
\\[0.4em]
$^{1}$Arizona State University
\quad
$^{2}$Luma AI
\quad
$^{3}$UC Davis
}

\newcommand{\method}[0]{\textsc{VisAnalog}}
\newif\ifshowcomments
\showcommentstrue
\ifshowcomments
  \newcommand{\ben}[1]{\textcolor{purple}{[Ben: #1]}}
  \newcommand{\zhaonan}[1]{\textcolor{blue}{[Zhaonan: #1]}}
\else
  \newcommand{\ben}[1]{}
  \newcommand{\zhaonan}[1]{}
  \newcommand{\marko}[1]{}
\fi

\begin{document}
\maketitle
\begingroup
\renewcommand{\thefootnote}{\ensuremath{\dagger}}
\footnotetext{Work done while at UC Davis.}
\endgroup

\begin{abstract}
A useful test of visual concept learning is not just whether a model can recognize a concept in a single image, but whether it can preserve and manipulate concept-level properties under transformation and transfer them to new scenes. We introduce \method{}, a controlled suite for this setting on natural images. Each example instantiates $A\!:\!B::C\!:\,?$: images $B$ and a hidden target image $D$ are produced by applying the same deterministic transformation sequence to source images $A$ and $C$. Given $A$, $B$, and $C$, a model must answer a multiple-choice question about $D$. The benchmark contains 617 human-validated questions spanning one- to four-step transformations such as zoom, quadrant swap, rotation, flip, and hue rotation. Across strong proprietary and open-source VLMs, end-to-end accuracy is substantially lower than oracle accuracy when $D$ is directly shown, and degrades sharply as transformation depth increases, while human performance remains near the ceiling. A program-conditioned evaluation further separates failures of relation inference from failures of transformation application, showing that inferring the visual relation from $A \rightarrow B$ is the dominant bottleneck, with additional application errors emerging on harder multi-step cases. The dataset is publicly available at \url{https://huggingface.co/datasets/zli99/VisAnalog}. 

% These results suggest that current VLMs are much better at reading visible evidence than at compositionally using visual concepts under relational transformation.
\end{abstract}

\begin{figure}
    \centering
    \includegraphics[width=1\linewidth, clip, trim={0.5cm 1.5cm 18.5cm 0cm}]{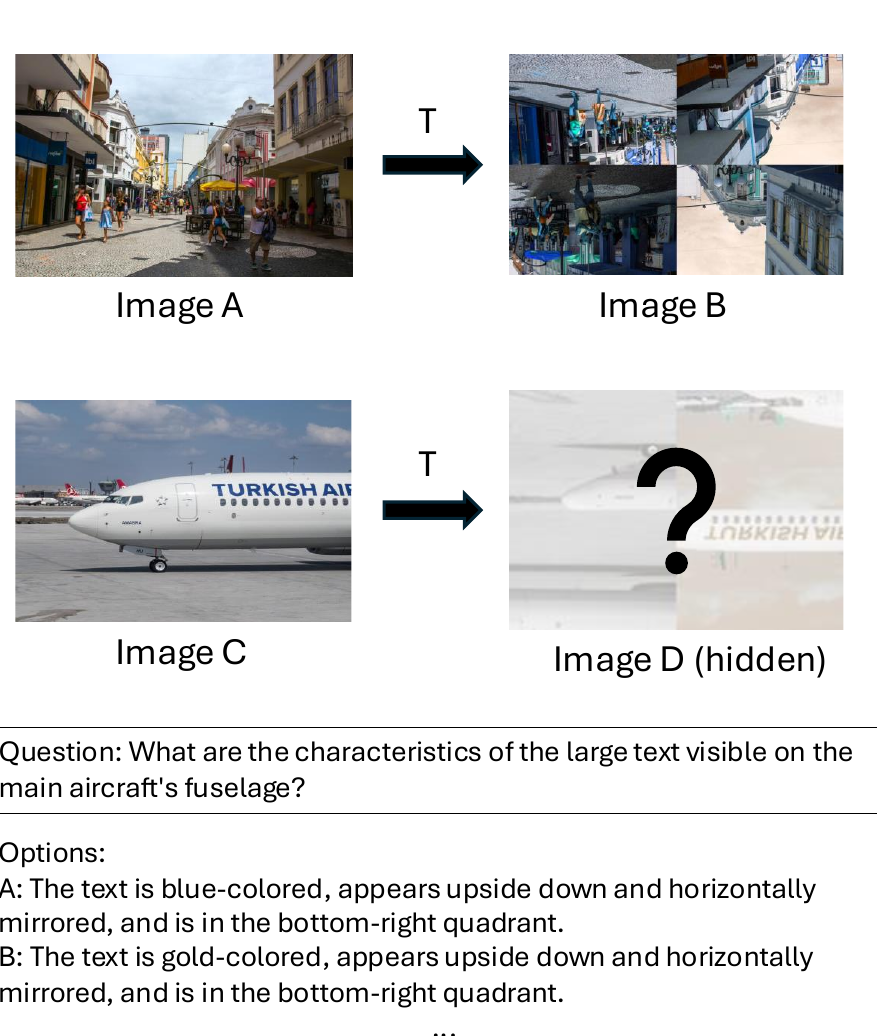}
    \caption{An example of the visual analogy questions for our benchmark. In these questions, source images $A$ and $C$ undergo the same transformation $T$, a sequence of image-editing operations, to become images B and D, respectively. During evaluation, the model is provided with images $A$, $B$, and $C$, and is tasked with answering a question about the hidden image $D$. To solve the question, the model must infer the abstract visual relationships between $A$ and $B$, then apply them to $C$ to mentally visualize $D$. The question assesses specific visual details in $D$ that depend on accurately inferring the transformed image. In this example, the question focuses on the color, orientation, and position of the text.}
    \label{fig:example_question}
\end{figure}

\section{Introduction}
% \label{sec:intro}

% A central goal of visual concept learning is to build representations that remain structured and usable beyond one-shot recognition.
% If a multimodal model has learned meaningful visual concepts, it should not only recognize them in a single image, but also track how visual properties and relations change under transformation and transfer those changes to new instances.
% This capability matters for visual reasoning, scene understanding, and controllable generation, where success depends on preserving and manipulating visual information rather than merely describing what is directly visible.

% We study this capability through a controlled visual analogy task on natural images.
% Each example instantiates an analogy $A \vcentcolon B \dblcolon C \vcentcolon ?$, where $B$ and a hidden target image $D$ are produced by applying the same deterministic transformation sequence to two source images $A$ and $C$.
% The model receives $A$, $B$, and $C$, but not $D$, and must answer a question about $D$.
% To succeed, it must infer the relation expressed by $A \rightarrow B$, transfer that relation to $C$, and reason about the resulting hidden image.
% Our goal is to test whether visual concepts are represented in a form that supports manipulation and compositional transfer. Across strong proprietary and open-source VLMs, performance is much higher when the true target image $D$ is directly shown than when the model must infer it, and end-to-end accuracy drops sharply as transformation depth increases while humans remain near ceiling.

\label{sec:intro}

A central goal of visual concept learning is to understand whether models can use visual concepts and relations beyond one-shot recognition.
If a multimodal model has learned meaningful visual concepts, it should not only recognize them in a single image, but also track how visual properties and relations change under transformation and transfer those changes to new instances.
This capability matters for visual reasoning, scene understanding, and controllable generation, where success depends on preserving and manipulating visual information rather than merely describing what is directly visible.
We study this capability through a controlled visual analogy task on natural images.
Each example instantiates an analogy $A \vcentcolon B \dblcolon C \vcentcolon ?$, where $B$ and a hidden target image $D$ are produced by applying the same deterministic transformation sequence to two source images $A$ and $C$.
The model receives $A$, $B$, and $C$, but not $D$, and must answer a question about $D$.
To succeed, it must infer the relation expressed by $A \rightarrow B$, transfer that relation to $C$, and reason about the resulting hidden image.
We use \emph{visual concepts} to refer to controllable image properties and \emph{visual relations} to refer to deterministic transformations over those properties. \method{} tests whether models can use such concepts and relations in a way that supports manipulation, composition, and transfer, rather than only recognizing them in a single image. Unlike ARC- or RAVEN-style analogy benchmarks built on synthetic patterns, our setting uses natural images, requiring models to identify and track these relations in realistic scenes while preserving precise control over the underlying transformation. Across strong proprietary and open-source VLMs, performance is substantially higher when the true target image $D$ is directly visible than when it must be inferred, and end-to-end accuracy declines sharply as transformation depth increases, even while human performance remains near ceiling.

\noindent\textbf{Contributions.} We make three contributions: (1) we introduce \method{}, a controlled benchmark for testing concept-level visual transfer under transformation on natural images; (2) we show that
strong VLMs struggle on this task, especially under multi-step
composition; and (3) we introduce a program-conditioned diagnostic
that makes the benchmark more interpretable by separating failures of
relation inference from failures of transformation application.

\section{Related Work}
\label{sec:related}

\noindent\textbf{Vision-centric evaluation of VLMs.}
Vision benchmarks for multimodal models span several settings.
Knowledge-centric VQA benchmarks test whether models can identify scene content and combine it with external or world knowledge \cite{goyal2017making,marino2019ok,schwenk2022okvqa,wang2017fvqa,shah2019kvqa}.
Relational and perception-focused benchmarks instead emphasize properties that must be read from the image itself, including spatial relations, orientation, and perceptual judgments \cite{kamath2023s,fu2024blink,stogiannidis2025mind}.
These works show that modern VLMs often struggle even when the relevant evidence is directly present in the input image.
Our benchmark is complementary: rather than asking about visible evidence alone, we ask models to reason about a \emph{hidden transformed target} derived from an observed image pair.

\noindent\textbf{Analogical reasoning and relational transfer.}
Analogical reasoning has long been studied as the transfer of relational structure rather than surface similarity \cite{gentner1983structure}, with links to visuospatial simulation and mental rotation in cognitive science \cite{krawczyk2012relational,shepard1971mental,richland2010analogical}.
In language models, analogical ability has been shown to emerge in some settings, but remains brittle when structure must be induced and transferred rather than recalled \cite{webb2023emergent,hu2023incontext,ye-etal-2024-analobench}.
Visual analogy benchmarks such as ARC, ConceptARC, and RAVEN similarly probe abstract relational transfer, but typically do so in synthetic or symbolic domains \cite{chollet2019measure,moskvichev2023conceptarc,mitchell2023comparing,zhang2019raven}.
We view \method{} as complementary to these benchmarks: instead of abstract symbolic patterns, we study controlled transformation-based transfer in natural images.

\noindent\textbf{Multimodal models and concept use under transformation.}
Recent unified multimodal models increasingly support both visual understanding and generation within a single architecture \cite{chen2025janus,xie2024show,deng2025emerging,bai2023qwen,yang2025qwen3}.
At the same time, recent evaluations continue to find major weaknesses in abstract, spatial, and transformation-based reasoning \cite{camposampiero2025can,zhang2024far,liang2025rover,ye2025blink,yilmaz2025voila}.
Our work differs from open-ended generation benchmarks and broad capability surveys in two ways.
First, we use a deterministic transformation family, which makes the underlying relation explicit and interpretable.
Second, we focus on whether visual information is processed in a way that supports compositional transfer and downstream reasoning.
In this sense, \method{} is best viewed not as a general benchmark of visual reasoning, but as a controlled probe of how multimodal models use visual concepts under relational transformation.

\section{\method{}: Task and Construction}
\label{sec:method}

\noindent\textbf{Task definition.}
We study a narrow form of visual analogy: \emph{visually grounded relational transfer}.
Each example instantiates an analogy of the form
$A \vcentcolon B \dblcolon C \vcentcolon ?$, where $B$ and a hidden target image $D$ are obtained by applying the same transformation sequence $T$ to two source images $A$ and $C$, respectively.
At test time, the model is given $A$, $B$, and $C$, but not $D$, and must answer a multiple-choice question about $D$.
To succeed, it must infer the relation expressed by $A \rightarrow B$, transfer that relation to $C$, and reason about the resulting hidden image.

We view this setup as a \emph{controlled concept probe} rather than a comprehensive benchmark of visual reasoning.
The goal is to test whether models can preserve and manipulate concept-level visual properties and relations---such as color, orientation, and spatial position---under controlled transformations in natural scenes.

\noindent\textbf{Controlled transformation family.}
We restrict $T$ to a fixed family of deterministic image edits:
(i) centered zoom (an 80\% crop resized back to the original resolution),
(ii) quadrant swap (exchanging two tiles in a $2\times2$ grid),
(iii) counter-clockwise rotation ($90^\circ$, $180^\circ$, or $270^\circ$),
(iv) horizontal or vertical flip, and
(v) hue rotation ($90^\circ$, $180^\circ$, or $270^\circ$).
Any example may contain a subset of these operations, with each operation used at most once.
When multiple operations are present, they are applied in a fixed order.
This keeps the task narrow but well specified: the underlying transformation is explicit, ambiguity from edit ordering is reduced, and difficulty can be controlled by the number of composed steps.

\noindent\textbf{Source images and instance generation.}
We construct examples from natural images sampled from SA-1B~\cite{kirillov2023segment}, favoring images with sufficient resolution and scene complexity.
For each instance, we sample two source images $A$ and $C$, generate a random transformation sequence $T$, and produce
$B=T(A)$ and $D=T(C)$ using PIL.
The use of natural images places these controlled transformations in visually rich scenes, while the synthetic edit family keeps the task programmatically precise.

% \noindent\textbf{Question generation.}
% For each pair $(C, D)$, we generate one multiple-choice question about $D$.
% Questions are written to target visual consequences of the transformation sequence, rather than generic scene semantics.
% In practice, the questions focus on concept-level properties and relations that change under transformation, such as object or text orientation, relative position, visible region, or color.
% Distractors are designed to reflect plausible partial failures, such as omitting one step or misapplying the transformation order.
% This design encourages success to depend on correctly transferring the visual relation, rather than on language priors alone. \zhaonan{mention the questions are proposed by LLM}
\noindent\textbf{Question generation.}
For each pair $(C, D)$, we generate one multiple-choice question about $D$.
Questions are written to target visual consequences of the transformation sequence.
% In practice, the questions focus on concept-level properties and relations that change under transformation, such as object or text orientation, relative position, visible region, or color.
The question, ground-truth answer, and distractor options are proposed by Gemini 2.5 Pro, conditioned on $C$, $D$, and the ground-truth transformation sequence.
Distractors are designed to reflect plausible partial failures, such as omitting one step or misapplying the transformation order.
This design ensures success depends on correctly transferring the visual relation, rather than on language priors alone.

% \noindent\textbf{Quality control.}
% We apply a two-stage filtering process.
% First, annotators verify that the transformation relating $A$ and $B$ is uniquely identifiable, removing ambiguous cases such as near-symmetries or repeated quadrants.
% Second, annotators verify that the question has a unique correct answer given accurate reasoning about $D$.
% In later iterations, these two checks were performed separately for more focused review, and we retain only items accepted by two annotators on both criteria.
% The final benchmark contains 617 questions spanning one- to four-step transformation sequences.
% \noindent\textbf{Quality control.}
% We apply a two-stage filtering process.
% First, annotators verify that the transformation relating $A$ and $B$ is uniquely identifiable, removing ambiguous cases such as near-symmetries or repeated quadrants.
% Second, annotators verify that the question has a unique correct answer given accurate reasoning about $D$.
% We retain only items accepted by two annotators on both criteria.
% To further ensure answerability and reduce potential annotation errors, we apply a post-processing step using two open-source VLMs, Qwen2.5-VL-72B and Llama-3.2-90B-Vision: each model answers the oracle VQA setting (question plus oracle $D$), and we keep only items solved by at least one of them.
% The final benchmark contains 617 questions spanning one- to four-step transformation sequences.

\noindent\textbf{Quality control.}
We apply a two-stage filtering process.
First, annotators verify that the transformation relating $A$ and $B$
is uniquely identifiable, removing ambiguous cases such as
near-symmetries or repeated quadrants. Second, annotators verify
that the question has a unique correct answer given accurate
reasoning about $D$. We retain only items accepted by two annotators
on both criteria. To reduce residual annotation errors in
answerability, we then apply a lightweight oracle-solvability check
with two open-source VLMs, Qwen2.5-VL-72B and
Llama-3.2-90B-Vision, using the question together with oracle $D$,
and keep only items solved by at least one model. We use this step
only to remove clearly problematic items, not to argue that the
benchmark is easy or saturated. The final benchmark contains
617 questions spanning one- to four-step sequences.

% \noindent\textbf{Evaluation settings.}
% Our main setting is \textbf{end-to-end analogy solving}: the model receives $A$, $B$, and $C$, and answers a question about hidden $D$.
% To contextualize performance, we also report:
% (1) an \textbf{oracle VQA} setting, where the model is given the ground-truth $D$;
% and (2) a \textbf{wrong-image control}, where the same question is asked using $C$ instead of $D$.
% The oracle setting measures whether the question is answerable given correct visual evidence, while the control checks that success depends on reasoning about the transformed target rather than on superficial cues or generic priors.

\section{Experiments}
\label{sec:experiments}

\noindent\textbf{Evaluation settings.}
We report three settings.
\textbf{End-to-end analogy} is the main task: the model receives $A$, $B$, and $C$, and answers a multiple-choice question about hidden $D$.
\textbf{Oracle VQA} provides the ground-truth target image $D$ directly and measures whether the question is answerable given correct visual evidence.
\textbf{Wrong-image control} replaces $D$ with $C$ and tests whether success can be explained by generic priors or superficial cues instead of reasoning about the transformed target.

% \noindent\textbf{Model selection.}
% Our original evaluation covered a broad range of proprietary and open-source VLMs.
% For the workshop version, we report a representative subset in the main text and defer the larger model suite to the supplement.
% We retain strong flagship closed models (GPT-o3, Gemini-2.5-Flash, Gemini-2.5-Pro) to show that the phenomenon is not limited to weaker baselines, and a small set of strong open models (Qwen2.5-VL-72B, InternVL3.5-38B-Instruct, Qwen3-8B-Instruct) to anchor the new decomposition experiment.
% We omit models whose interpretation is confounded by output truncation or whose small scale adds little beyond the same qualitative trend.

\noindent\textbf{Model selection.}
We report a representative set of strong proprietary and open-source VLMs, spanning frontier closed models (GPT-o3, Gemini-2.5-Flash, Gemini-2.5-Pro) and strong open models (Qwen2.5-VL-72B, InternVL3.5-38B-Instruct, Qwen3-VL-8B-Instruct). 
% This set is sufficient to show that the observed trend is not confined to a single model family.

\noindent\textbf{Human baseline.}
We treat the human result as a sanity-check estimate
rather than a high-precision benchmark. Following the benchmark
protocol, we randomly sample 40 questions (10 per step count), ask
two annotators to solve them independently with access to $A$, $B$,
$C$, and the transformation family, and adjudicate disagreements to
form consensus labels.

\subsection{Main benchmark results}

\begin{table}[t]
\centering
\small
\setlength{\tabcolsep}{4pt}
\begin{tabular}{lcccc}
\toprule
Model & Step1 & Step2 & Step3 & Step4 \\
\midrule
Human & 100.0 & 90.0 & 90.0 & 90.0 \\
GPT-o3 & 77.4 & 66.0 & 49.0 & 30.7 \\
Gemini-2.5-Flash & 67.3 & 59.7 & 51.6 & 24.0 \\
Gemini-2.5-Pro & 71.2 & 49.7 & 48.4 & 29.3 \\
Qwen2.5-VL-72B & 51.8 & 39.6 & 43.3 & 52.0 \\
InternVL3.5-38B-Instruct & 51.3 & 32.1 & 21.7 & 26.7 \\
Qwen3-VL-8B-Instruct & 58.4 & 34.0 & 29.3 & 30.7 \\
\bottomrule
\end{tabular}
\caption{End-to-end analogy accuracy (\%) on \method{}.}
\label{tab:main_results}
\end{table}

\noindent\textbf{Strong models still struggle with transformation-based visual analogy.}
Table~\ref{tab:main_results} reports representative end-to-end results.
Humans remain near ceiling across all step counts, while strong closed and open models are substantially worse and, for most models, degrade as the number of composed transformations increases.
This trend is especially clear for GPT-o3 and the Gemini models: for example, GPT-o3 drops from 77.4\% on 1-step questions to 30.7\% on 4-step questions, while humans remain at 90--100\% across all four bins.
Overall, the table shows that transferring a visually specified relation and then reasoning about the hidden result is much harder than ordinary single-image question answering.

% A small caveat is that the decline is not strictly monotonic for every model.
% In particular, Qwen2.5-VL-72B shows a rebound at Step~4 rather than a smooth downward trend.
% We do not over-interpret this pattern.
% Part of it may reflect dataset construction: during post-filtering, we retained questions that were solvable by at least one of two large open-source VLMs---Qwen2.5-VL-72B or Llama3.2-90B---when given the oracle target image $D$, which may mildly favor models with similar oracle-VQA strengths.
% Regardless, even this strongest open model remains far below human accuracy and still exhibits a substantial gap between end-to-end analogy and oracle grounding, so the qualitative conclusion is unchanged.

\begin{figure}[t]
    \centering
    \includegraphics[width=\linewidth]{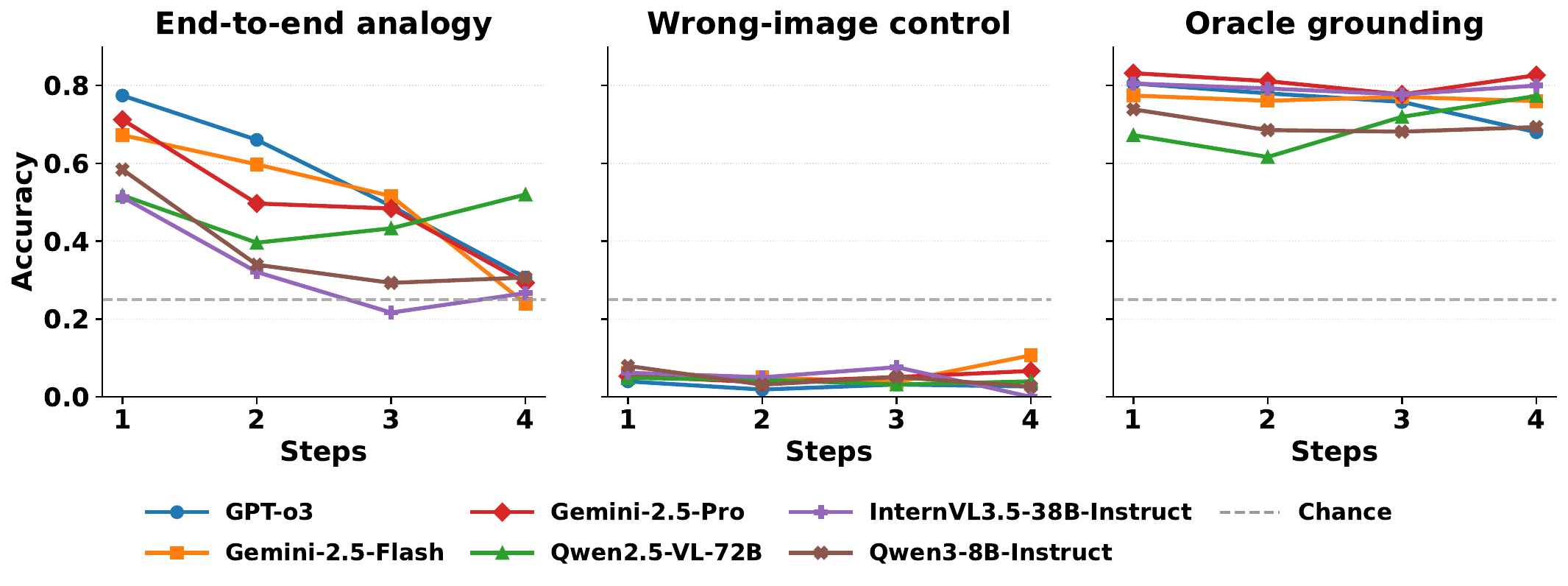}
    \caption{Comparison of end-to-end analogy solving, wrong-image control, and oracle VQA.
    Models perform much better when the true target image $D$ is visible than when they must infer it, while the wrong-image control stays near or below chance.
    Dashed lines indicate 25\% random-guessing accuracy.}
    \label{fig:plot2_controls}
\end{figure}

\noindent\textbf{The benchmark requires transferring transformed visual concepts, not guessing from the source image.}
Figure~\ref{fig:plot2_controls} provides the key control.
Questions are explicitly written to target properties and relations that are \emph{true in $D$}, must follow from the transformation sequence, and are \emph{not answerable from $C$ alone}; distractors are designed to reflect plausible misapplications such as omitted steps or wrong order.
Consistent with that design, models perform below chance in the wrong-image control, where they are asked the question using the source image $C$ instead of the hidden target $D$.
In contrast, when the ground-truth target image $D$ is directly provided, oracle accuracy remains high across models, often above 80\%.
Together, these controls show that the task is not solved by generic priors or by reading off concepts already visible in the source image.
Instead, success requires \emph{transferring} concept-level visual information through the inferred transformation and then reasoning over the resulting hidden target.

\noindent\textbf{The main difficulty is the analogy gap, and it widens with composition.}
Taken together, Table~\ref{tab:main_results} and Figure~\ref{fig:plot2_controls} isolate the core difficulty of \method{}.
The questions are answerable when correct visual evidence is available, but not from the untransformed source image, which means the remaining gap is specifically about analogical transfer of visual relations.
Moreover, the separation between oracle grounding and end-to-end analogy grows as the transformation chain becomes longer, indicating that errors accumulate when models must preserve and manipulate transformed visual concepts across multiple steps.
The central empirical finding of our benchmark is a growing composition gap: current VLMs remain much stronger at answering from direct visual evidence than at inferring, composing, and applying multi-step transformations.
% This is the central empirical finding of our benchmark: 
% current VLMs are substantially better at answering from direct visual evidence than at internally executing multi-step concept-preserving visual transformations over natural images.

% \noindent\textbf{Oracle grounding is much easier than inferred grounding.}
% To verify that the questions themselves are solvable, we compare end-to-end analogy solving against oracle VQA with the true target image $D$ visible.
% Across models, oracle VQA is substantially easier, showing that the difficulty does not come from the question format alone.
% Conversely, in the wrong-image control where the model sees $C$ instead of $D$, performance drops below random-guessing levels.
% Together, these controls indicate that \method{} depends on access---direct or inferred---to the transformed target image rather than on generic priors alone.

% \begin{figure}[t]
%     \centering
%     % Replace with your final plot.
%     \fbox{\rule{0pt}{1.6in}\rule{0.95\linewidth}{0pt}}
%     \caption{\textbf{Placeholder.} Accuracy by transformation depth for oracle VQA ($D$ visible), end-to-end analogy ($A,B,C$), and the wrong-image control ($C$ in place of $D$). The original evaluation shows the consistent pattern oracle VQA $>$ end-to-end analogy $\gg$ wrong-image control.}
%     \label{fig:oracle_control}
% \end{figure}

\subsection{Diagnosing inference and application errors}
\label{sec:exp_diag}

\noindent\textbf{Motivation.}
Figure~\ref{fig:plot2_controls} measures end-to-end performance on the analogy reasoning task, requiring the model to both infer the composition of visual concepts and relations from $A$ to $B$ and apply it to $C$. To better understand this capability, we isolate two failure modes, separating errors in inferring the analogy from errors in applying it. We add a program-conditioned setting in which the model receives $C$ together with the exact ground-truth transformation program $T$, and must answer the same question about hidden $D$.
This removes relation inference while preserving the need to apply a multi-step transformation and reason over the resulting target.

\noindent\textbf{Metrics.}
We report program-conditioned accuracy $\mathrm{Acc}(C,T)$ and two signed gaps:
\[
\begin{aligned}
\Delta_{\mathrm{inf}} &= \mathrm{Acc}(C,T)-\mathrm{Acc}(A,B,C), \\
\Delta_{\mathrm{app}} &= \mathrm{Acc}(D)-\mathrm{Acc}(C,T).
\end{aligned}
\]
where $\mathrm{Acc}(A,B,C)$ is end-to-end analogy accuracy and $\mathrm{Acc}(D)$ is oracle VQA accuracy.
A large $\Delta_{\mathrm{inf}}$ indicates that inferring the relation from $A \!\rightarrow\! B$ is a major bottleneck; a positive $\Delta_{\mathrm{app}}$ indicates residual difficulty in applying the known transformation and using the transformed visual concepts for question answering.

\begin{table}[t]
\centering
\small
\setlength{\tabcolsep}{4pt}
\begin{tabular}{llccccc}
\toprule
Model & Metric & S1 & S2 & S3 & S4 \\
\midrule
\multirow{3}{*}{InternVL3.5-38B}
& $\mathrm{Acc}(C,T)$            & 70.8 & 70.4 & 62.4 & 53.3 \\
& $\Delta_{\mathrm{inf}}$        & 19.5 & 38.4 & 40.8 & 26.7 \\
& $\Delta_{\mathrm{app}}$        &  9.7 &  8.8 & 15.3 & 26.7 \\
\midrule
\multirow{3}{*}{Qwen3-VL-8B}
& $\mathrm{Acc}(C,T)$            & 72.1 & 67.9 & 61.8 & 52.0 \\
& $\Delta_{\mathrm{inf}}$        & 13.7 & 34.0 & 32.5 & 21.3 \\
& $\Delta_{\mathrm{app}}$        &  1.8 &  0.6 &  6.4 & 17.3 \\
\bottomrule
\end{tabular}
\caption{Program-conditioned evaluation. $\mathrm{Acc}(C, T)$ gives accuracy when the model is provided with image $C$ and the exact transformation program $T$. $\Delta_{\mathrm{inf}}$ measures the gain from removing relation inference, and $\Delta_{\mathrm{app}}$ measures the residual gap to oracle grounding. Results are reported separately for each step count.}
\label{tab:diagnostic_gaps}
\end{table}

\noindent\textbf{Findings.}
Across both models, the inference gap is large at every step and is generally larger beyond Step 1, indicating that relation inference is a major source of error in \method{}. This suggests that a substantial portion of the end-to-end failure arises before the model ever reasons about the hidden target: it often fails to recover how visual properties and relations change from $A$ to $B$. At the same time, the application gap is also consistently positive and becomes more pronounced as the composition grows longer. For InternVL3.5-38B-Instruct, $\Delta_{\mathrm{app}}$ rises from 9.7 points at Step 1 to 26.7 points at Step 4, indicating substantial residual difficulty even when the correct transformation is provided. Qwen3-VL-8B-Instruct shows a smaller residual gap, but it too increases with depth, from near zero on Steps 1--2 to 17.3 points on Step 4. Taken together, these trends show that both error sources are meaningful, and that both become more visible as the sequence of visual concept changes grows. More broadly, the negative correlation with step number suggests that current models do not yet exhibit robust, compositional mastery of visual concept recognition and transfer.
% \noindent\textbf{Caveat on Qwen2.5-VL-72B.}
% We exclude Qwen2.5-VL-72B from the main diagnostic table because it was used during the model-based post-filtering stage that retained questions solvable by at least one large open-source VLM under oracle grounding, which may mildly favor this model. :contentReference[oaicite:3]{index=3}
% Its results follow the same broad pattern and are reported in the supplement.

\section{Conclusion}

We introduced \method{}, a diagnostic benchmark for evaluating whether VLMs can recover a visual relation from one pair of natural images and use it to reason about an unseen counterpart. Each example instantiates an analogy $A:B::C:?$, where the model must infer the edit sequence underlying $A \rightarrow B$, compose the corresponding operations on $C$, and answer a question about the latent target. Controlled evaluations confirm that the benchmark measures visually grounded relation identification and execution rather than single-image recognition or shortcut reasoning from the source image. Our main finding is a widening composition gap: as transformation depth increases, end-to-end accuracy falls substantially, revealing a weakness in current VLMs' ability to maintain and manipulate visual concepts across multi-step edits. 
% Program-conditioned experiments further show that this gap arises from both relation-inference failures and residual errors in applying known transformations, with both sources of error becoming more salient as compositions grow longer.
Program-conditioned experiments show that this gap reflects both relation-inference failures and residual errors in applying known transformations, both worsening with composition length.

\newpage
{
    \small
    \bibliographystyle{ieeenat_fullname}
    \bibliography{main}
}

\appendix
\section{Model Prompts}

\begin{promptbox}{Question Generation}
========================
1) Context and Inputs
========================
You are a question generator that writes exactly one diagnostic multiple-choice question (MCQ) about the target image D.

You are given:
- C: the source image.
- D: the ground-truth target image produced by applying a sequence of transformations to C.
- Sigma = [tau_1, tau_2, ..., tau_k]: an ordered list of transformations mapping C -> D.

Transformation types are for your internal reasoning only; do NOT mention them in the question or options:
- Operational: pixel-space edits, e.g., rotation, center crop, flip, hue shift, quadrant swaps.

Evaluation usage: Later, a solver model will not see D. It will receive A, B, and C, infer A -> B, apply the inferred transformation(s) to C to imagine D, and then answer your question about D. Your question must probe visual details of D that are consequences of the transformation(s), thereby testing whether the solver can simulate transformations mentally.

========================
2) Objective
========================
1. Understand the visual consequences that distinguish D from C via Sigma.
2. Write one self-contained MCQ about D that:
   - Makes sense on its own.
   - Never mentions C, D, "transformation," "analogy," or step names.
   - Requires correct simulation of the full sequence of transformations to answer.
   - Is not answerable from C alone, generic priors, or an incorrect visualization of the target image.
3. Provide four options, labeled A, B, C, and D, with exactly one correct answer.
4. Make each distractor a plausible outcome of a specific mis-simulation: omitted step, wrong order, or wrong interpretation.
5. Provide an explanation proving why the correct option is uniquely true in D and diagnosing each distractor.

================================
3) Hard Leak-Prevention Rules
================================
Never reveal, hint at, or imply any of the following in the question or options:
- The existence of transformations, Sigma, step types, or operation names.
- Any verbs or phrases that imply change or causality, e.g., "becomes," "turned," "after," "before," "now," "transformed," "once rotated," "when aged," "if winter arrives," "gets flipped," or "shifted."
- Any meta-language about the protocol, e.g., A/B/C/D, "analogy," "simulation," or "apply the transformation."
- Direct references to D, such as "in the final image."

Focus on neutral, stative facts about what is true in D: objects, attributes, spatial relations, and states. Do not ask "what changed"; only ask "what is."

==================================
4) Robustness and Diagnostic Power
==================================
- Not answerable from C alone or generic priors. The correct answer must hinge on the effects of the transformation sequence that produces D.
- Plausible failure modes. Write distractors that reflect realistic mis-visualizations, e.g., skipped steps, wrong order, or wrong magnitude, so the item challenges an unfaithful visualizer.
- Salient, stable consequences. Target robust, clearly visible outcomes that persist in a correct rendering of D; avoid tiny details or subtle, hairline differences.
- Parallel and balanced options. Keep the answer choices similar in length and style.

=============================
5) Output Format
=============================
Print exactly this JSON object, with no extra text and no code fences:

{
  "rationale": "<brief reasoning: which consequences of C -> D are targeted; why solving requires the entire sequence; how each distractor challenges an incorrect or inaccurate visualizer>",
  "question": "<one single-sentence MCQ about D; no mention of C, D, transformations, or analogy>",
  "options": ["<A>", "<B>", "<C>", "<D>"],
  "explanation": "<why the chosen option is uniquely correct for D only when all steps are applied>",
  "answer": "<A|B|C|D>"
}
\end{promptbox}

\begin{promptbox}{Analogy Reasoning}
You are a Visual Analogical Reasoner.

Task
- Solve the analogy A : B :: C : D.
- You are given three images: A, B, and C.
- Infer the minimal transformation T that maps A -> B, then mentally apply T to C to imagine D.
- Then answer a fine-grained multiple-choice question about the imagined D.

Input
- image_A, image_B, image_C
- question: a fine-grained query about the imagined D
- choices: four options labeled "A", "B", "C", and "D"

Output
- Let's think step by step, and then output only one capital letter in a LaTeX box, e.g., \boxed{A}.

Each transformed image is produced by up to 4 non-repeating operations applied in a fixed order: a centered zoom, a swap of two tiles in a 2x2 quadrant grid, a counter-clockwise rotation of 90, 180, or 270 degrees, a horizontal or vertical flip, and a hue rotation of 90, 180, or 270 degrees. Any subset of these operations may be used, but whenever an operation is included, it appears in this order.

Question:
[Question]

Options:
[Options]

Image A:
[Image A]

Image B:
[Image B]

Image C:
[Image C]
\end{promptbox}

\begin{promptbox}{Standalone VQA Evaluation}
You are a Visual Question Answering model.

Task
- Answer a multiple-choice question about the image.

Input
- image
- question: a query about the image
- choices: four options labeled "A", "B", "C", and "D"

Output
- Let's think step by step, and then output only one capital letter in a LaTeX box, e.g., \boxed{A}.

Question:
[Question]

Options:
[Options]

[Input Image]
\end{promptbox}

\end{document}